\documentclass{article}
\usepackage{spconf,amsmath,graphicx}

\usepackage{amsfonts}
\usepackage{amsthm}
\usepackage{amssymb}
\usepackage{amsmath}
\usepackage{enumitem}
\usepackage{multirow}
\usepackage{comment}
\usepackage{subcaption}
\usepackage{booktabs}

\title{On the Efficiency of Integrating Self-supervised Learning and Meta-learning for User-defined Few-shot Keyword Spotting}
\name{Wei-Tsung Kao$^1$$^\star$, Yuan-Kuei Wu$^1$$^\star$, Chia-Ping 
Chen$^2$, Zhi-Sheng Chen$^2$, Yu-Pao Tsai$^2$, Hung-Yi Lee$^1$\thanks{$^\star$The two first authors made equal contributions.}}
\address{
  $^1$Graduate Institute of Communication Engineering, National Taiwan University\\
  $^2$intelliGo Technology inc.\\
  r09942067@ntu.edu.tw, f07942100@ntu.edu.tw, ailsa.chen@intelli-go.com,\\ cs.chen@intelli-go.com,
yptsai@gmail.com,
hungyilee@ntu.edu.tw
 }

\begin{document}
%
\maketitle
\begin{abstract}
  User-defined keyword spotting is a task to detect new spoken terms defined by users. This can be viewed as a few-shot learning problem since it is unreasonable for users to define their desired keywords by providing many examples. To solve this problem, previous works try to incorporate self-supervised learning models or apply meta-learning algorithms. But it is unclear whether self-supervised learning and meta-learning are complementary and which combination of the two types of approaches is most effective for few-shot keyword discovery. In this work, we systematically study these questions by utilizing various self-supervised learning models and combining them with a wide variety of meta-learning algorithms. Our result shows that HuBERT combined with Matching network achieves the best result and is robust to the changes of few-shot examples.
\end{abstract}
\begin{keywords}
Keyword Spotting, Few-shot Learning, Self-supervised Learning, Meta-learning
\end{keywords}
\section{Introduction}


Keyword-spotting (KWS) is a task to detect specific words in speech streams, which is an essential function in recent smart devices for users to access remotely by speech.
To obtain an accurate KWS system, a common approach is that manufacturers pre-define the keywords of their products, and then collect large-scale datasets to train KWS models.
This is practical but not optimal due to limited personalization. That is, these kinds of KWS models can not allow user-defined keywords.

For user-defined keywords, large datasets are not available since we cannot ask the users to provide many examples. So it can be treated as a few-shot learning problem. The approaches proposed by previous works fall into two categories: 

\begin{itemize}[wide, labelindent=0pt]
    \item Transfer from labeled data of other keywords: 
    Chen \cite{chen20j_interspeech} applies Model Agnostic Meta-Learning (MAML)~\cite{finn2017model} to learn a better initialization for fine-tuning on the few keyword examples. 
    Parnami \cite{parnami2020few} use Prototypical network~\cite{snell2017prototypical} to learn an encoder that can cluster the embeddings of the examples of the same keywords. 
    But Chen  do not obtain satisfying accuracy, and Parnami  only conducts simpler 2-classes and 4-classes classification experiments. 
    Lin \cite{lin2020training} train the encoder of their KWS model to classify the keyword groups on 200M clips from YouTube and synthesized speech from a text-to-speech model.  
    Huang \cite{9414156} train an embedding model on LibriSpeech~\cite{7178964} by softtriple loss~\cite{qian2019softtriple}, which also clusters the embeddings of the same keywords while allowing multiple centers in each keyword.
    Awasthi \cite{awasthi21_interspeech} train a multi-class keyword classification model on LibriSpeech as their encoder and show better generalization ability under few-shot learning. 
    Mazumder \cite{mazumder21_interspeech} train a multi-class multilignual keyword classification model with EfficientNet's structure~\cite{tan2019efficientnet} as the encoder on Common Voice~\cite{ardila2020common} to solve multilingual few-shot KWS problem. 
    Nonetheless, preparing large-scale KWS datasets usually requires audios, transcription, and a forced aligner, which increases the cost. Or it would rely on an external text-to-speech system to synthesize data, which could suffer from domain mismatch. 
    
    \item Utilize unlabeled data: Approaches in the first category aim to learn a better encoder for KWS. From this viewpoint, using large-scale labeled data may not be necessary. Self-supervised learning (SSL) is an effective method to learn feature extractors from unlabeled data. 
    Seo \cite{9427206} incorporate Wav2Vec 2.0~\cite{baevski2020wav2vec}, a SSL model, into their KWS models. 
    However, since the authors focus on the performance with the full dataset available, the performance of 1-shot learning is poor.
\end{itemize}

\begin{figure*}[t]
  \centering
  \includegraphics[width=.88\linewidth]{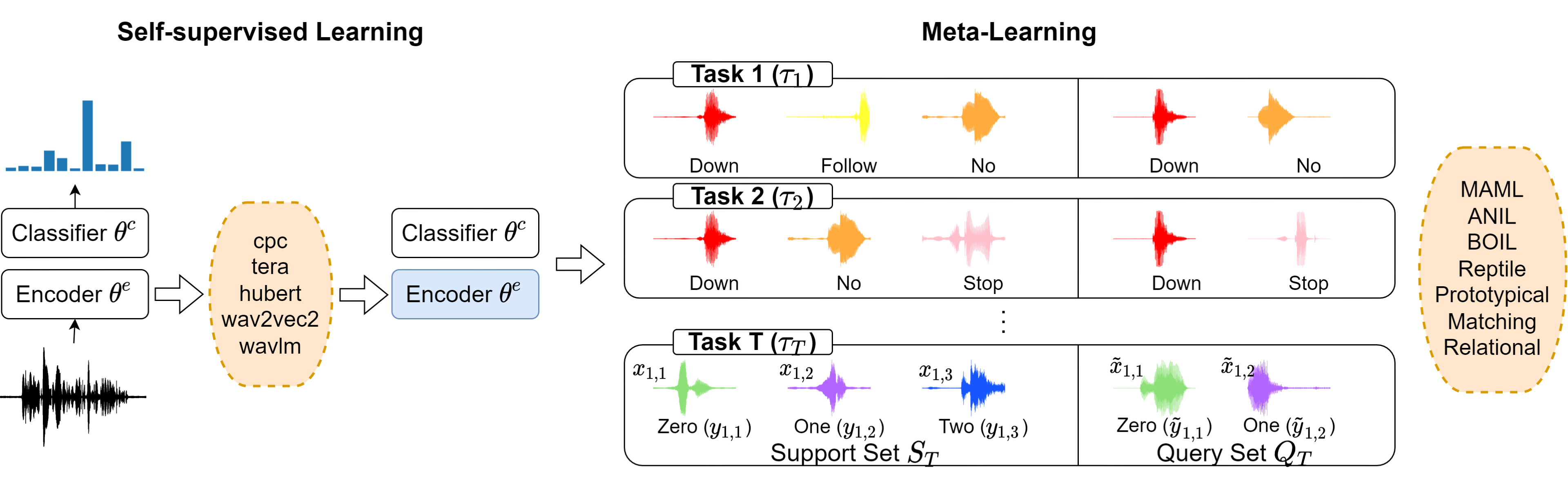}
  \caption{The pipeline of our methods. First, we will decide whether to initial the encoder with the pre-trained self-supervised learning model and fix the parameters or not. Then, we will train our model using meta-learning algorithms. The white module blocks represent the randomly initialized parameters and the blue encoder block refer to the model pre-trained from SSL.}
  \label{fig:pipeline}
\end{figure*}

Are the two types of approaches above complementary?
It has been found that the integration of PASE+~\cite{ravanelli2020multi} and meta-learning including Prototypical network and MetaOptNet~\cite{lee2019meta} improves the  keyword spotting performance~\cite{mittal20_interspeech}.
But to our best knowledge, it is still unclear whether the effects of SSL and meta-learning are additive in general and independent of specific choices of SSL models or meta-learning algorithms.  
What kinds of combinations is the most appropriate for few-shot KWS is also uninvestigated.
In this paper, we systematically study the combination of state-of-the-art SSL models and typical meta-learning algorithms. Our contributions include:
\begin{itemize}
    \item We compare several widely used SSL models to answer which pre-trained model is the best for few-shot KWS.
    \item Training the SSL models by seven meta-learning algorithms, we shed light on the effectiveness of combining the pre-training and meta-learning approaches.
\end{itemize}

\section{Methods}

\label{sec:method}
The overall pipeline is illustrated in Fig. \ref{fig:pipeline}. 
Our KWS system is composed of an encoder and a classifier whose parameters are denoted as $\theta^e$ and $\theta^c$, respectively. Given an utterance, the encoder outputs a sequence of vectors as the representations and inputs them to the classifier. 
Then the classifier outputs the posterior probability of each keyword based on the extracted features.
The encoder can be a pre-trained SSL model or randomly initialized parameters. 
We train the KWS model with meta-learning algorithms, so the model can fast adapt to new keyword sets. 
The learnable parameters during adaptation can be either the whole model or only the classifier. Furthermore, the encoder can be fixed if initialized with the pre-trained SSL weights.

\subsection{Self-supervised learning}

Self-supervised learning (SSL) is a widely used to utilize unlabeled data. 
We explore five SSL models as the encoder in our experiments: CPC~\cite{oord2018representation}, TERA~\cite{liu2020tera}, Wav2Vec2~\cite{baevski2020wav2vec}, HuBERT~\cite{hsu2021hubert}, and WavLM~\cite{chen2021wavlm}. 
CPC pre-trains an LSTM model that predicts the latent representations of future speech frames by the representations of current speech frames. 
TERA masks the spectrogram of the input speech along time and frequency and pre-trains a transformer model to reconstruct the unmasked speech. 
Wav2Vec2 masks part of input representations and pre-trains a transformer model to reconstruct the representations by loss similar to CPC. 
HuBERT clusters the MFCC frames of the pre-training data, masks input representations like Wav2Vec2, and predicts the cluster id at the masked positions. 
WavLM uses learning targets similar to HuBERT but extends the SSL pre-training to simulated noisy data. These pre-trained models can extract informative representations better than traditional speech features such as MFCC and significantly improve the performance of speech processing tasks such as ASR~\cite{baevski2020effectiveness}.

In this work, we use the 3-layer TERA model and the 12-layer Wav2Vec2 model, HuBERT model, and WavLM model as our encoder.
TERA takes the mel-spectrogram of an utterance $x$ as input, and the other four models take the raw waveform of $x$ as input. 
Each layer of these SSL models outputs a sequence of vectors $\{h_t\in\mathbb{R}^d\}_{t=1}^{T}$ for each time frame $t$.
Here we ignore the notation of layer for simplicity.
We average $\{h_t\}_{t=1}^{T}$ to get a single representation $\Bar{h}$ for $x$.
Then $\Bar{h}$ from each layer is weighted summed by trainable weights, and input to a keyword classifier to decide which keyword $x$ contains.

\subsection{Meta-learning}
\label{sec:method_meta}

Meta-learning is designed for training models that can be easily transferred to a set of new tasks with few examples.
Let $\theta=\theta^e \cup \theta^c$ be the model parameters of the KWS model and $f_\theta$ be the model parameterized by $\theta$.
For user-defined KWS, meta-learning trains $f_\theta$ on a meta-train dataset $\hat{D}$ consisting of several known keywords and tests it on a meta-test dataset $D$ composed of new keywords. Specifically, $\hat{D}=\{\hat{\tau_i}\}$ is a set of keyword spotting tasks $\hat{\tau_i}$. Each task $\hat{\tau}_i=\{\hat{S}_i, \hat{Q}_i\}$ is a $N$-way-$K$-shot problem. That is, it requires detecting $N$ different keywords. For each keyword $w$ has only $K$ training examples $\hat{S}_i=\{(x_{j,w}, y_{j,w})| 1\leq j \leq K, 1\leq w\leq N\}$, where $x_{j,w}$ is a waveform of an utterance contains keyword $w$, $y_{j,w}$ is a label. 
$\hat{S}_i$ is referred to as support set. $\hat{Q}_i = \{(\tilde{x}_{j,w}, \tilde{y}_{j,w})\}$ is the set of testing examples called query set. There is no any assumption for the number of examples in $\hat{Q}_i$. $D=\{\tau_i\}$ is similar except that the labels in $\hat{Q}_i$ would be used for updating model parameters in meta-train, while labels of $Q_i$ in $\tau_i$ are only used for evaluation. 

We investigate seven meta-learning algorithms: MAML, ANIL~\cite{raghu2019rapid}, BOIL~\cite{oh2020boil}, Reptile~\cite{nichol2018first}, Prototypical network~\cite{snell2017prototypical}, Relational network~\cite{sung2018learning}, and Matching network~\cite{vinyals2016matching}.
The first four algorithms are optimization-based methods, which requires $f_{\theta^c}: \mathbb{R}^d \rightarrow \mathbb{R}^N$. 
We do not include MetaOptNet due to convergence and stability issues of the convex problem solver.
And the other three algorithms are metric-based methods, in which $f_{\theta^c}: \mathbb{R}^d \rightarrow \mathbb{R}^n$ is an embedding function. $n$ depends on algorithms but does not necessarily equal $N$. 
Metric-based methods usually use a non-parametric classifier $\mathcal{C}$ such as k-NN on top of $f_{\theta^c}$.
We incorporate SSL models into meta-learning by loading their parameters as the initialization of $\theta^e$.

\begin{figure}[t]
  \centering
  \includegraphics[width=.88\linewidth]{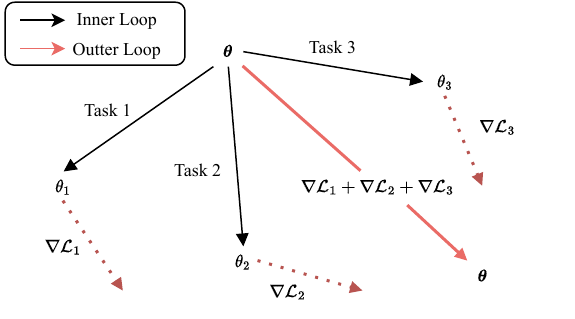}
  \caption{The illustration of MAML and its variants.}
  \label{fig:MAML}
\end{figure}

\subsubsection{Optimization-based methods}

In deep learning, we use gradient descent to find a local minimum for a given function. However, it usually requires a significant amount of data and many optimization steps to converge. Optimization-based meta-learning intends to find initial model parameters that can rapidly adapt to user-defined KWS models, where $\theta^e$ is the parameters of the encoder, and $\theta^c$ is the ones of the classifier. 

There are two steps in optimization methods, the inner loop and the outer loop (Fig \ref{fig:MAML}). 
The inner loop adapts $\theta$ to task-specific model $\hat{\theta}_i$ on the support set $\hat{S}_{i}$ by gradient descent, which we also refer to as the meta-optimization step.
The task-specific models $\theta_i$ are updated from $\theta$ using gradient descent with the cross-entropy loss $\mathcal{L}$ and learning rate $\alpha$, formulated as
\begin{align}
\mathcal{L}_{\hat{S}_i}(f_{\theta}) &= - \mathbb{E}_{(x, y) \in \hat{S}_i} y \log f_{\theta} (x)\\
\theta_i &\leftarrow \theta - \alpha \nabla_{\theta} \mathcal{L}_{\hat{S}_i}(f_{\theta})   \label{eq:meta-update}
\end{align}
We can apply gradient descent multiple times. After acquiring the task-specific models  $\theta_i$, the outer loop updates the parameter $\theta$ with respect to the cross-entropy loss calculated by the task-specific models and the query set.
\begin{align}
    \theta \leftarrow \theta - \beta \sum_{i}  \nabla_{\theta} \mathcal{L}_{\hat{Q}_i}(f_{\theta_i})
    \label{eq:outer-loop}
\end{align}
MAML and its variants ANIL and BOIL differ mainly in their inner loop. 
MAML updates the whole model in the inner loop;
that is, the parameters $\theta$ to be trained include the parameters of the encoder $\theta^e$ plus the classifier $\theta^c$ in the eq \ref{eq:meta-update}. 
ANIL points out that $\theta^e$ already contains high-quality representations through the outer loop optimization. In consequence, $\theta^e$ is fixed in the inner loop;
on the contrary, 
BOIL leverages the representation changes, fixes $\theta^c$, and updates $\theta^e$ in the inner loop. 

\begin{figure}[t]
  \centering
  \includegraphics[width=.88\linewidth]{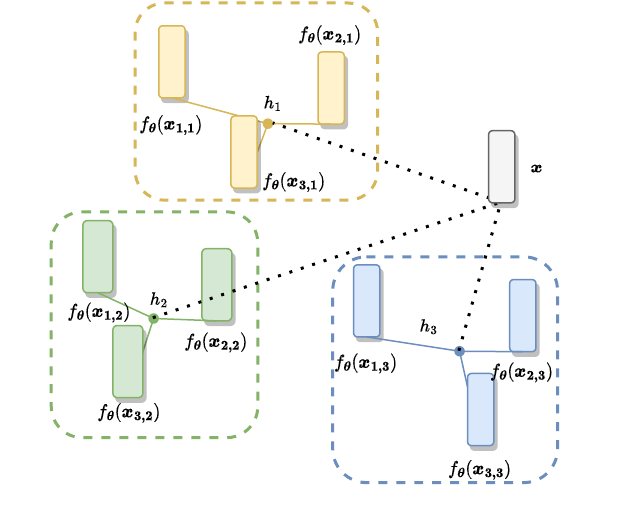}
  \caption{Prototypical Network}
  \label{fig:proto}
\end{figure}
 
To find the optimal initialized parameters $\hat{\theta}$, MAML, ANIL, and BOIL apply gradient descent. They calculated the gradient with respect to the loss mentioned in eq \ref{eq:outer-loop}. 
However, this requires computing the second derivative, which is computationally expensive. 
So we adopt the first-order approximation proposed in MAML.
A simplified version is called First-Order MAML (FOMAML), which omits the second derivatives. The update is formulated as
\begin{align}
    \label{eq:fomaml}
    \theta \leftarrow \theta - \beta \sum_{i}  \nabla_{\theta_i} \mathcal{L}_{\hat{Q}_i}(f_{\theta_i})
\end{align}
Where $\beta$ is the learning rate, the inner loop of Reptile is similar to MAML, and the difference is mainly in the outer loop. Instead of calculating the gradient of the loss with respect to the task-specific model, Reptile replaces the calculation of gradient simply by the difference between $\hat{\theta}_i$ and $\theta$. 

\begin{align}
    \theta \leftarrow \theta - \beta \sum_{i} (\theta - \theta_i)
\end{align}

\subsubsection{Metric-based methods}
Metric-based meta-learning aims to learn an embedding model such that embeddings of uttreances of the same keyword should be as close as possible.
\begin{itemize}[wide, labelindent=0pt]
    \item Prototypical network (Fig. \ref{fig:proto}) selects k-NN for $\mathcal{C}$. For each keyword $w$, it averages $\{f_\theta(x_{j,w})\}_{j=1}^K$ to be the keyword embedding $h_w$ of $w$. 
    During training, $L_2$ distances between $f_\theta(\tilde{x}_{j,w})$ and $h_w$'s are logits and trained by CE.
    During testing, the model first calculates the keyword embeddings by the support set. Given an utterance $x$, the model classifies it as the keyword whose embedding is the closest to $f_\theta(x)$.
    \item Relational network averages $\{f_{\theta^e}(x_{j,w})\}_{j=1}^K$ as $h_w$, concatenates $f_{\theta^e}(\tilde{x}_{j,w})$ to $h_w$s, and use $f_{\theta^c}$ to output a scalar (relation score) for each $w$ without using non-parametric $\mathcal{C}$. 
    It is trained by mean square error to make the relation score of keyword $\tilde{y}_{j,w}$ to 1, and 0, otherwise.
    \item Matching network also chooses $\mathcal{C}$ to be k-NN, while it applies attention mechanism in $f_{\theta^c}$ to encode $\{f_{\theta^e}(x_{j,w})\}_{j=1}^K$ for all $w$ and each $f_{\theta^e}(\tilde{x}_{j,w})$ into $NK$ support embeddings and one query embedding. 
    Therefore, different from prototypical network and relational network, these embeddings depend on the whole support set and the query example.
    $L_2$ distance between the query embedding and each support embedding is transformed to probability of the corresponding keyword by softmax. 
    The probability of the same keyword is summed and trained by cross-entropy. 
    
\end{itemize}

\subsection{Classifier and training details}

We apply a learnable weighted sum to the representations from every layer of the encoder and add a 4-layer ReLU DNN classifier on top of the encoder, except that we use a 1-layer transformer in Matching Network. 
The numbers of parameters of the 4-layer DNN and the 1-layer transformer are both 7.1M, which are 7.5\% of the one of the 12-layer HuBERT.
We use Adam optimizer for metric-based methods and outer loop of optimization-based methods and SGD as inner loop optimizer. 
The learning rate of SGD is set to $5\times 10^{-2}$, which is better for MAML among $\{10^{-1}, 5\times 10^{-2}, 10^{-2}, 10^{-4}, 10^{-5}\}$. 
The learning rate of Adam is set to $10^{-4}$. We adapt 5 steps during the meta-train and 20 steps during the meta-test. 
The meta-batch size is 4 in all the experiments. 
We train all the models to either convergence or at most 20 epochs. 

\section{Results on Google Speech Commands}

\subsection{Dataset}

In our meta-learning experiments, each keyword spotting task is a 12-way-$K$-shot classification problem. The classes consist of 10 keywords classes, "unknown" class, and "silence" class. $K$ different utterances are available for each class. 
In the experiments, $K$ is set to 1 or 5.
The utterances in the "silence" class are fixed-length clips from background noises. 
Following previous works, we use Google Speech Commands V2 dataset (Speech Commands)~\cite{speechcommandsv2}, which consists of 35 keywords and a total of 105,829 utterances. 
We select 5 keywords as the "Unknown" keywords, 20 keywords for meta-train, and the remaining 10 keywords for meta-test. Keyword utterances used in meta-train and meta-test are disjoint.
However, there are only 6 background noise utterances in the dataset. 
Clipping from only these utterances to form the silence class for meta-training and meta-testing could make the task too simple. 
Therefore, we use noises from WHAM! noise dataset~\cite{wichern2019wham} instead of the ones in Speech Commands. 
WHAM! consists of 23K noise utterances collected from real-world urban environments. We follow the splits in the dataset and make the utterances used in meta-train and meta-test disjoint, too.
During meta-train, we randomly sample 1000 tasks in each epoch. 
For meta-test, we sample 1000 tasks once and fix the tasks. 
So all models are tested on the same tasks.

\subsection{Baselines}

Our dataset is harder than original Speech Commands. 
Thus, we do not make an apples-to-apples comparison between our experiment results and the scores in previous works. 
Instead, we compare these meta-learning algorithms with three baselines: (1) Transfer-v1 (Trans-1): This baseline is a 20-way classification model trained on the 20 keywords previously used in meta-train. The model structure is the same as the one used in MAML except for the last liner layer. During testing, we replace the last linear layer in the classifier with a randomly initialized 12-class linear layer and fine-tune the models on the $K$-shot examples of the testing keywords. (2) Transfer-v2~\cite{awasthi21_interspeech} (Trans-2): We train HuBERT on LibriSpeech by the task proposed in~\cite{awasthi21_interspeech} and fine-tune the model on the $K$-shot examples of the testing keywords. (3) \textit{scratch}: We train randomly initialized models with the same structure as HuBERT by meta-learning.
The learning rate setup is the same as optimization-based meta-learning.

\begin{table*}[t]
    \centering
    \caption{Accuracy of meta-learning combined with different SSL models and baselines. "fine-tune" means all parameters are trainable. "fix-encoder" means the SSL models are frozen when trained on the downstream task.}
    \begin{tabular}{llcccc|ccc|c|c}
        \toprule
        & SSL & MAML & ANIL & BOIL & Reptile & Prototypical & Matching & Relational & Trans-1 & Trans-2\\
        \midrule
        & CPC & 31.64 & 46.18 & 21.08 & 27.79 & 46.40 & \textbf{46.98} & 40.81 & 8.58 & \\
        \multirow{3}{*}{\parbox{41pt}{\centering 1-shot\\ fine-tune}} & TERA & 44.66 & 39.93 & 43.97 & 37.84 & 48.12 & \textbf{53.62} & 42.16 & 44.88 & \\
        & HuBERT & 50.00 & 63.13 & 38.53 & 53.78 & 67.99 & \textbf{70.39} & 49.34 & 63.33 & 41.12\\ 
        & Wav2Vec2 & 53.10 & 56.60 & 53.47 & 45.10 & 63.39 & 64.82 & 38.97 & \textbf{65.71} & \\
        & WavLM & 39.12 & 53.88 & 46.34 & 38.81 & 69.90 & \textbf{76.16} & 42.83 & 58.26 & \\
        \midrule
        & CPC & 33.97 & - & - & 23.48 & 39.73 & 41.63 & 35.71 & \textbf{47.69} & \\
        \multirow{3}{*}{\parbox{41pt}{\centering 1-shot\\ fix-encoder}} & TERA & 41.55 & - & - & 27.90 & 43.00 & \textbf{48.18} & 37.91 & 45.45 & \\
        & HuBERT & 61.43 & - & - & 47.34 & 70.03 & \textbf{79.30} & 64.18 & 66.37 & 56.58\\
        & Wav2Vec2 & 57.41 & - & - & 35.04 & 56.69 & \textbf{71.07} & 57.99 & 66.5 & \\ 
        & WavLM & 63.84 & - & - & 33.75 & 55.51 & \textbf{75.27} & 64.12 & 59.61 & \\
        \midrule
        & CPC & 32.02 & 58.49 & 21.68 & 52.05 & \textbf{67.90} & 64.55 & 59.39 & 9.06 & \\
        \multirow{3}{*}{\parbox{41pt}{\centering 5-shot\\ fine-tune}} & TERA & 52.89 & 68.39 & 69.92 & 69.59 & \textbf{75.40} & 73.93 & 58.15 & 66.76 & \\
        & HuBERT & 65.26 & 83.18 & 79.85 & 83.95 & 85.88 & \textbf{88.98} & 56.21 & 84.93 & 79.95\\
        & Wav2Vec2 & 60.58 & 78.76 & 70.84 & 82.45 & 80.49 & \textbf{86.47} & 52.89 & 84.82 & \\ 
        & WavLM & 80.72 & 82.26 & 82.35 & 81.24 & 78.51 & \textbf{87.30} & 58.35 & 81.52 & \\
        \midrule
        & CPC & 30.88 & - & - & 35.60 & 56.98 & \textbf{58.32} & 51.61 & 49.62 & \\
        \multirow{3}{*}{\parbox{41pt}{\centering 5-shot\\ fix-encoder}} & TERA & 45.56 & - & - & 44.67 & 60.55 & 62.71 & 50.93 & \textbf{66.6} & \\
        & HuBERT & 70.80 & - & - & 38.02 & 85.84 & \textbf{90.86} & 73.60 & 85.03 & 78.42\\
        & Wav2Vec2 & 54.53 & - & - & 53.95 & 82.68 & \textbf{85.52} & 76.00 & 84.88 & \\
        & WavLM & 70.24 & - & - & 49.02 & 83.06 & \textbf{86.39} & 67.75 & 81.16 & \\
        \bottomrule
    \end{tabular}
    \label{tab:1shot}
\end{table*}

\subsection{Comparision between algorithms}

\begin{table}[t]
    \centering
    \caption{Standard deviation of accuracy on testing tasks.}
    \begin{tabular}{lccc}
        \toprule
         & ANIL & Matching & Trans-1 \\
        \midrule
        1-shot fine-tune  & 6.82 & 6.24 & 12.99\\
        5-shot fine-tune & 3.23 & 2.58 & 3.92 \\
        \bottomrule
    \end{tabular}
    \label{tab:std}
\end{table}

Table~\ref{tab:1shot} shows the average accuracy of different SSL models, meta-learning algorithms, and baselines. 
We do not apply ANIL and BOIL when the encoder is fixed because under this setting, ANIL is the same as MAML, and we can not perform inner loop updates for BOIL.
For meta-learning, metric-based methods outperform popular optimization-based methods in general. 
Among metric-based algorithms, Matching network reaches better performance in 1-shot and 5-shot learning. 
It also reaches the best performance among all experiments when combined with HuBERT. 
Relational network is consistently worse than the other two algorithms in this category. 
For optimization-based algorithms, MAML is the best for 1-shot learning, while Reptile outperforms MAML under the 5-shot learning setup. 
ANIL obtains accuracy comparable to MAML for 1-shot learning and Reptile for 5-shot learning.
So we can consider that ANIL is the most compatible with SSL models in this category.

Compared with transfer-v1, Matching network achieves better results, while ANIL's performance is usually below the one of transfer-v1. 
So in terms of average performance, we do not benefit from meta-learning if we apply optimization-based methods. 
However, when investigating the standard deviation of the performance across different testing tasks and support examples in Table~\ref{tab:std}, we observe that ANIL is more robust to support set changes than transfer-v1, especially under the 1-shot learning scenario.
And Matching network is not only the best on average but also the most robust algorithm.
This feature is crucial for user-defined KWS applications where few-shot examples provided by different users could vary a lot.
Finally, comparing transfer-v1 and transfer-v2, we can realize that labeling the corpus used in SSL pre-training and training KWS classifiers on it could be inefficient due to poor generalizability. 

\subsection{Comparison between SSL models}
\label{sec:main_result}

In the last sub-section, we have shown that metric-based methods consistently perform better than optimization-based methods. 
Therefore, here we only compare different SSL models when trained by metric-based meta-learning. 
Although WavLM performs better on SUPERB~\cite{yang21c_interspeech} benchmark which does not consider few-shot learning, HuBERT takes a leading position in both 1-shot and 5-shot learning. Fixing HuBERT encoder largely improves the performance, which implies that HuBERT's representations are suitable for meta-learning. This property is preferable in terms of reducing training costs. Different from HuBERT, fine-tuning the encoder sometimes enhances Wav2Vec2's and WavLM's results such as using Prototypical network for 1-shot learning and Matching network for 5-shot learning. 
The performance of CPC and TERA is disappointing. And fine-tuning these two encoders consistently outperform their fixed encoder counterparts. Thus, we conjecture that their model size limits the strength of representations.

\subsection{The synergy between SSL and meta-learning}

\begin{table}[t]
    \centering
    \caption{Comparison of accuracy between HuBERT and the trained from scratch models.}
    \begin{tabular}{lccc}
       \toprule
        & Prototypical & Matching & Relational \\
       \midrule
       1-shot HuBERT  & 67.99 & 70.39 & 49.34 \\
       1-shot scratch & 38.95 & 40.80 & 41.23\\
       \midrule
       5-shot HuBERT  & 85.88 & 88.98 & 56.21 \\
       5-shot scratch & 61.56 & 60.26 & 50.91 \\
       \bottomrule
    \end{tabular}
    \label{tab:scratch}
\end{table}

Now we have shown that SSL models trained by Matching network can surpass the ones without meta-learning for few-shot KWS. 
To show that SSL and meta-learning are synergistic, it remains to verify whether initializing from SSL models contributes to the performance or not. 
Hence, we try to train the KWS models from scratch by metric-based meta-learning and compare the results. In Table~\ref{tab:scratch}, there are large gaps between HuBERT and the trained from scratch models across all algorithms. The gaps remain even when number of examples increases. Consequently, both SSL and meta-learning contribute to the performance when applied at the same time. The effect of SSL and the effect of meta-learning are additive.

\begin{figure}[t]
    \centering
    \begin{subfigure}[t]{0.325\columnwidth}
        \centering
        \includegraphics[width=\columnwidth]{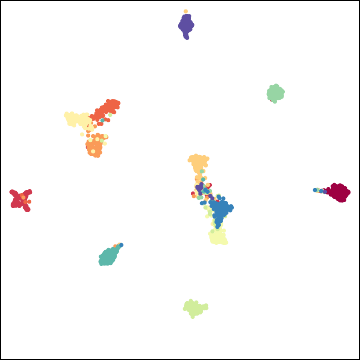}
        \caption{meta+SSL}
    \end{subfigure}
    \begin{subfigure}[t]{0.325\columnwidth}
        \centering
        \includegraphics[width=\columnwidth]{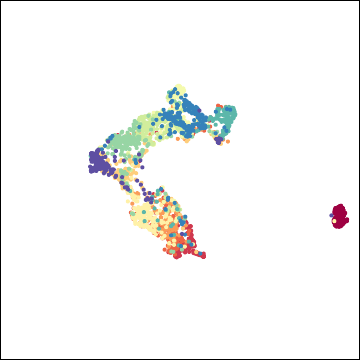}
        \caption{meta only}
    \end{subfigure}
    \begin{subfigure}[t]{0.325\columnwidth}
        \centering
        \includegraphics[width=\columnwidth]{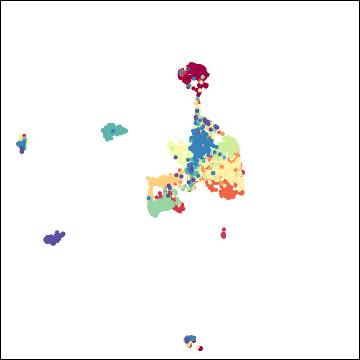}
        \caption{SSL only}
    \end{subfigure}
    \caption{PaCMAP visualization of (a) HuBERT+Matching network, (b) Matching network trained from scratch, and (c) HuBERT without any fine-tuning on the testing keywords. The colors stand for keywords. Each point represents one utterance.}
    \label{fig:pacmap}
\end{figure}

To elaborate on this phenomenon, we visualize the embedding space of the (unseen) keywords used in the testing tasks by PaCMAP~\cite{JMLR:v22:20-1061}. PaCMAP balances the global structure and the local structure information during projection and preserves more local information than tSNE~\cite{van2008visualizing}. We compare Matching network with HuBERT as the encoder (meta+SSL), Matching network without HuBERT (meta only), and the last layer representation of HuBERT (SSL only) without fine-tuning on the KWS dataset. 
The result is in Figure~\ref{fig:pacmap}, where different colors stand for different keywords. 
For the SSL-only model, a part of the keywords is clearly separated while some keywords overlap; for the meta-only model, there are no manifest clusters. Different keywords are just slightly distinguishable; For the meta+SSL model, the boundaries of different keywords are obvious. The points of the same keyword are concentrated. Therefore, SSL representations contain non-trivial information for KWS, which can enhance meta-learning and obtain more discriminative embeddings.




\section{Results on Common Voice dataset}

\begin{table}[t]
    \centering
    \caption{Results on Common Voice. The encoder is fixed when we apply meta-learning.}
    \begin{tabular}{lccc}
        \toprule
         & ANIL & Matching & Trans-1 \\
        \midrule
        1-shot & 63.62 & 80.41 & 22.77 \\
        5-shot & 77.32 & 90.3 & 24.33 \\
        \bottomrule
    \end{tabular}
    \label{tab:common_voice}
\end{table}

Finally, we perform an experiment on the Common Voice dataset to validate that our conclusion remains the same across different datasets. 
In this experiment, we use the single English word segments provided in the dataset, which contains 14 words and a total of 32726 utterances\footnote{We exclude 6567 utterances longer than 3 seconds.}.
Each task in this experiment is a 6-way-$K$-shot keyword spotting problem. 
The classes consist of 4 keywords, "unknown" class and "silence" class. 
We utilize eight keywords during meta-train, four during meta-test, and two keywords for "unknown."
Similar to the previous experiment, the utterances in the "silence" class come from WHAM!, and $K$ is 1 or 5. 
Notice that the original sampling rate of the utterances in Common Voice is 48kHz. 
We downsampled them to 16kHz to fit the sampling rate during SSL pre-training.

Here we conduct an experiment with the best model, HuBERT, and the better algorithms, ANIL, and matching network according to the previous experiment.
The results are in table~\ref{tab:common_voice}. 
The matching network outperforms Trans-1 baseline and ANIL, which is consistent with the experiment on Speech Commands.
Therefore, the effect of SSL and meta-learning are additive across different datasets.

\section{Conclusions}

In this paper, we systematically study the combination of self-supervised learning and meta-learning to solve user-defined keyword-spotting problems, which are usually few-shot learning problems. Our extensive experiments show that combining HuBERT and Matching network can achieve the best performance under 1-shot and 5-shot learning scenarios and be robust to the variation of different few-shot examples. Our analyses validate that the effect of self-supervised learning and the effect of meta-learning are additive, making the embeddings of unseen keywords more distinguishable. 

\bibliographystyle{IEEEbib}
\bibliography{mybib,refs}

\end{document}